\documentclass[10pt,twocolumn,letterpaper]{article}

\usepackage[pagenumbers]{cvpr} 

\usepackage{graphicx}
\usepackage{amsmath}
\usepackage{amssymb}
\usepackage{booktabs}
\usepackage{enumitem}
\usepackage{color}
\usepackage{array}

\usepackage[pagebackref,breaklinks,colorlinks]{hyperref}

\usepackage[capitalize]{cleveref}
\crefname{section}{Sec.}{Secs.}
\Crefname{section}{Section}{Sections}
\Crefname{table}{Table}{Tables}
\crefname{table}{Tab.}{Tabs.}

\def\cvprPaperID{1490} 
\def\confName{CVPR}
\def\confYear{2023}

\newcommand{\tablestyle}[2]{\setlength{\tabcolsep}{#1}\renewcommand{\arraystretch}{#2}\centering\footnotesize}

\begin{document}

\title{More comprehensive facial inversion for more effective expression recognition.}

\author{
Jiawei Mao$^\dag$
\quad Guangyi Zhao$^\dag$  \quad Yuanqi Chang  \quad Xuesong Yin{\thanks{Corresponding author.$^\dag$Equal contribution.}} \quad  Xiaogang Peng \quad Rui Xu  \\ 
School of Media and Design, Hangzhou Dianzi University, Hangzhou, China \qquad \\
{\tt\small\{jiaweima0,gyzhao,yuanqichang,yinxs,pengxiaogang,211330017\}@hdu.edu.cn }\\
}
\maketitle

\begin{abstract}
    Facial expression recognition (FER) plays a significant role in the ubiquitous application of computer vision. We revisit this problem with a new perspective on whether it can acquire useful representations that improve FER performance in the image generation process, and propose a novel generative method based on the image inversion mechanism for the FER task, termed Inversion FER (IFER). Particularly, we devise a novel Adversarial Style Inversion Transformer (ASIT) towards IFER to comprehensively extract features of generated facial images. In addition, ASIT is equipped with an image inversion discriminator that measures the cosine similarity of semantic features between source and generated images, constrained by a distribution alignment loss. Finally, we introduce a feature modulation module to fuse the structural code and latent codes from ASIT for the subsequent FER work. We extensively evaluate ASIT on facial datasets such as FFHQ and CelebA-HQ, showing that our approach achieves state-of-the-art facial inversion performance. IFER also achieves competitive results in facial expression recognition datasets such as RAF-DB, SFEW and AffectNet. The code and models are available at ~\url{https://github.com/Talented-Q/IFER-master}.
\end{abstract}

\section{Introduction}
\label{sec:intro}

\begin{figure}[t]
  \centering
   \includegraphics[width=1\linewidth]{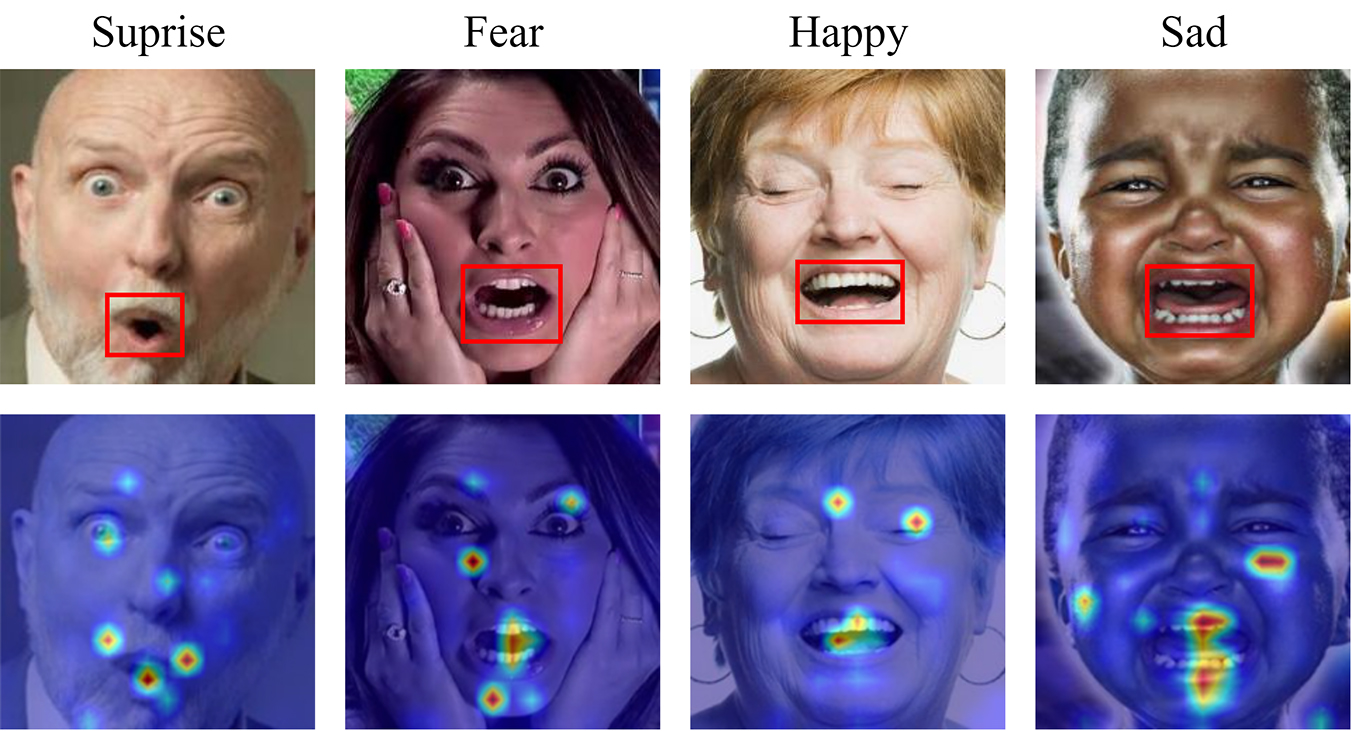}
   \caption{Attention visualizations on four types of facial expression examples. Row 1: Original images. Row 2: Attention visualizations of our ASIT algorithm. Our ASIT not only observes the red box areas, but also several areas related to expressions. There is no doubt that global scope observation can better help IFER to correctly analyze facial expressions. See Appendix {\color{red}C} for more attention visualization results.}
   \label{fig:1}
\end{figure}

\begin{figure*}
  \centering
  \includegraphics[width=1\linewidth]{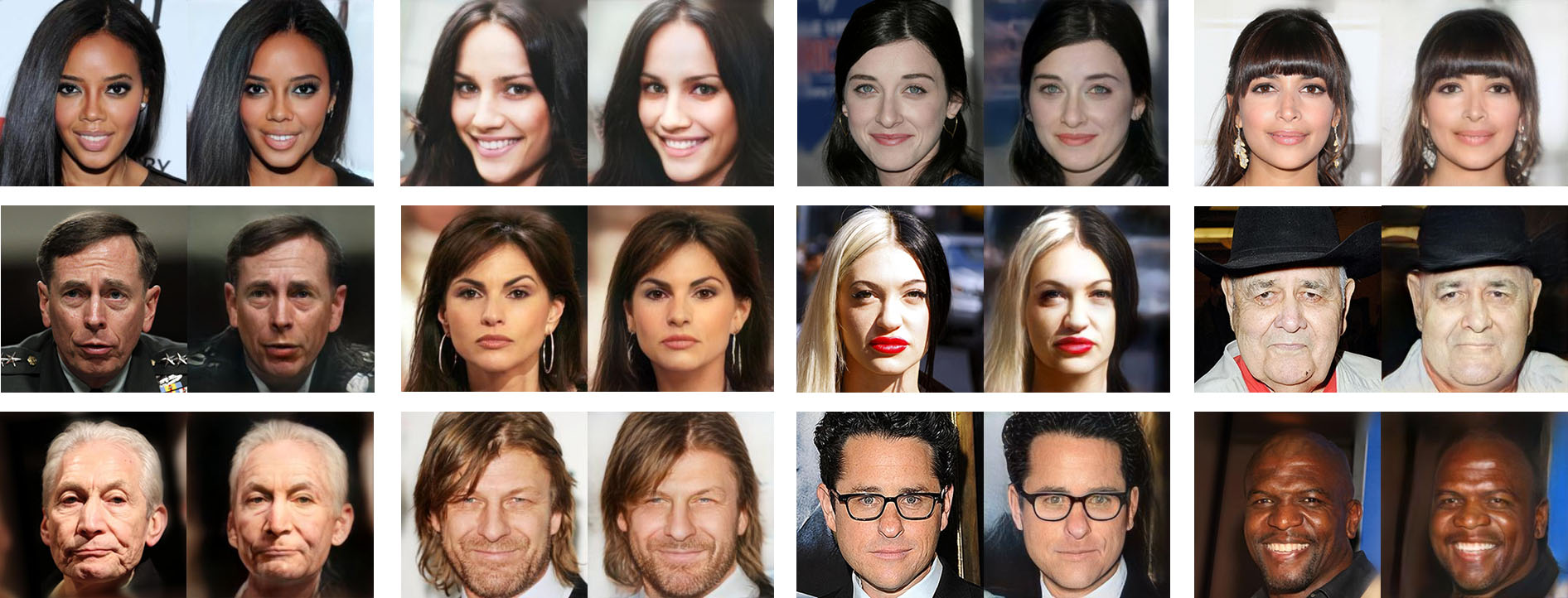}
    \caption{Our ASIT achieves the state-of-the-art performance of facial image inversion. Here, we show the inversion results of ASIT on the facial dataset CelebA-HQ\cite{bib24}. The left image is the source image and the right image is our inversion.}
    \label{fig:2}
  \hfill
\end{figure*}

The technique for Facial Expression Recognition (FER) plays an important role in human-computer interaction and human-human communication by recognizing facial information and analyzing human psychological emotions. It has great potential applications in the fields of psychology, intelligent robotics, intelligent surveillance, virtual reality and synthetic animation \cite{bib1,bib2,bib3,bib4,bib5,bib6,bib7}. Traditional FER approaches use manual features for expression recognition \cite{bib8,bib9,bib10}. With the rise of deep learning, many researchers attempt to exploit neural networks to solve the problem of expression recognition. For instance, Fan \etal \cite{bib11} deeply fuse recurrent neural networks and 3D convolutional networks to effectively solve the video expression recognition problem. In recent years, many studies have shown great promotion of facial features on the FER task. Ding \etal \cite{bib12} propose a FaceNet2ExpNet and incorporated facial domain knowledge for regularization during training. Benefiting from the advantages of the attention mechanism in Transformer, Xue \etal \cite{bib13} first propose a Transformer-based method called TransFER that learns the intrinsic relationship between different local patches via dropping one self-attention module randomly.

Different from prior methods, we treat this issue from a new perspective that better learns representations for FER during the process of facial expression generation. A rising technique in image generation called image inversion boosts learning the intrinsic structure by capturing the key facial information in the latent space \cite{bib14,bib15,bib16,bib17,bib21}.

Although the currently proposed image inversion models show excellent performance in a variety of tasks related to faces, few researchers focus on their performance on the FER problem. Furthermore, current image inversion models. \cite{bib14,bib15} tend to select only the middle layer features of the image for inversion and use constant noise as the initial input of the generator. \cite{bib18} propose a point that the initial input feature map of the generator can control the structural information of the image, and the highest feature map in the encoder contains rich semantics. We consider that it is useful to take advantage of the a prior knowledge embedded in image generation and learn the high-level semantic information associated with it. To this end, we propose a new expression recognition algorithm, called Inverted FER (IFER). Specifically, we first perform image inversion pre-training to understand the process of expression generation. In order to fully train the high-level semantic features in our image inversion model, we replace the constant-level noise in StyleGAN2 \cite{bib19} with it and name it as the structure code. Next, how to better fuse the latent codes obtained from the image inversion model with the structure code for expression recognition is a important issue in our IFER. In StyleGAN2, the latent codes are continuously adjusted to the constant noise at different levels by modulated convolution to finally generate a realistic image. So, we design a feature modulation module for our IFER. We choose modulated convolution as the fusion scheme for latent codes and the structure code. Since different levels of latent codes have different natures of influence on the structure code and higher levels of latent codes correspond better to the structure code, the feature modulation module uses Multi-Layer Perceptrons (MLPs) to adjust the weight of each latent code layer corresponding to the structure code. 

Simple fusion is not enough to present satisfactory results, while the image inversion model itself needs to be improved. Most of the current image inversion models use convolutional neural networks, which extract only local details of the face while paying little attention to the overall structure of the face. Actually, expressions are expressed through various parts of the face most of the time. As shown in \cref{fig:1}, if the network can only focus on a part of the face, it may result in incorrect expression understanding. Therefore, we propose an image inversion algorithm with a pure transformer architecture and call it the Adversarial Style Inversion Transformer (ASIT) to capture the global information of the image. In addition, the following improvements have been made to further enhance ASIT’s understanding of image composition: First, we design a discriminator model specifically for image inversion. It extracts high-level semantic information between pairs of images and calculates the cosine similarity as an output. Second, since image inversion is the inversion process of image generation, we consider that the corresponding layer features of the generator and our image inversion model should obey the same distribution. This can be explained by the fact that each layer of the encoder can observe the same level of features in the corresponding layer of the generator if they have the same distribution. This can help our ASIT further utilize the prior knowledge of GAN to better understand the image generation process. Therefore, we introduce the distribution alignment loss \cite{bib22} into ASIT. Third, we design a new local-to-global feature extraction branch by adjusting the window size of the window attention. \cref{fig:2} shows the facial inversion effect of ASIT, which achieves state-of-the-art results on the dataset.

\begin{figure*}[t]
  \centering
  \includegraphics[width=0.9\linewidth]{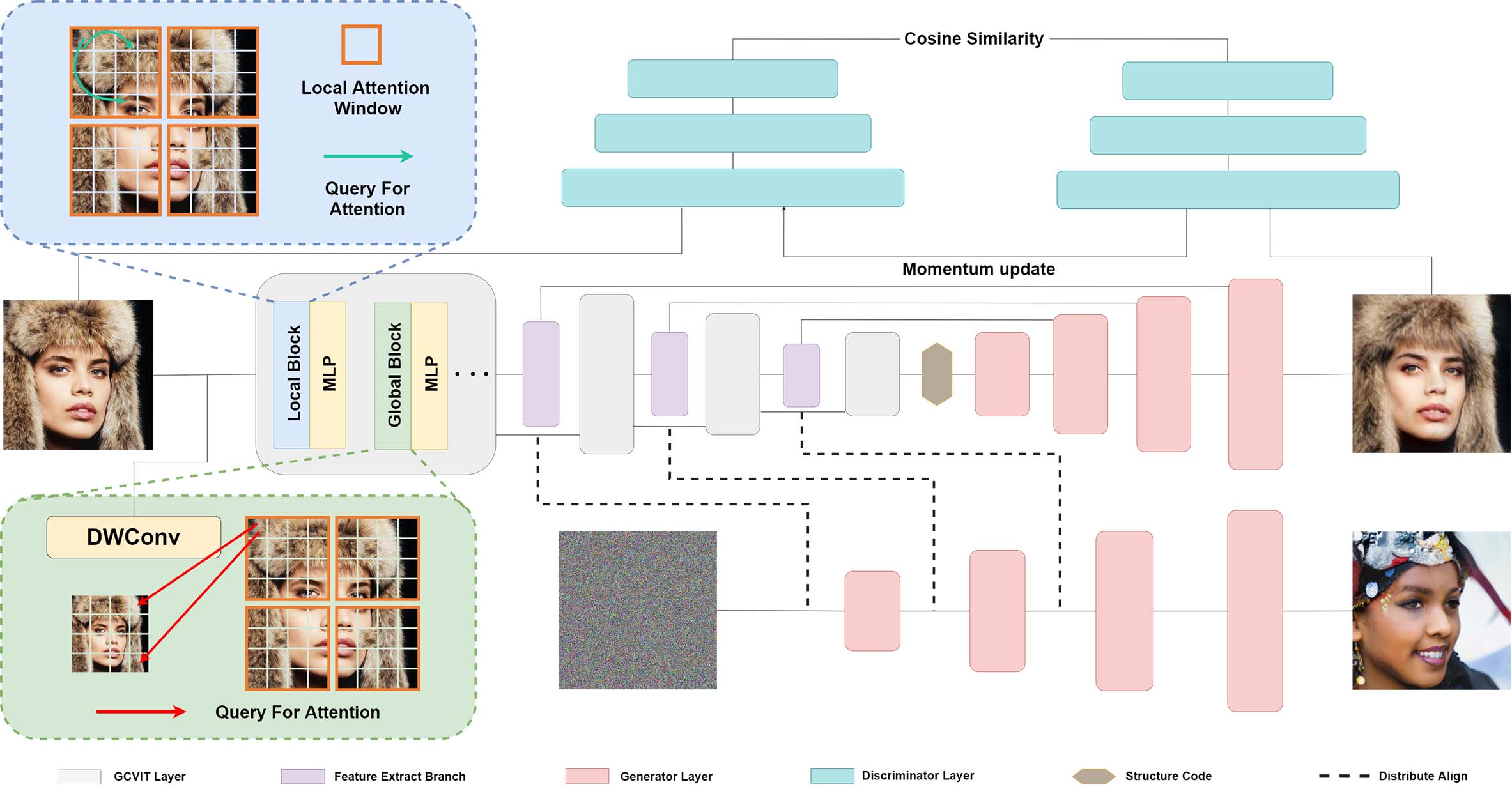}
    \caption{Our ASIT framework. It uses GC ViT\cite{bib20} as the base model, which contains global attention for long-range dependence and window attention for local information. Next, the latent codes of the image are extracted in a local-to-global manner through our feature extraction branch. In addition, we obtain the structure code of the image by projecting the highest-level features. Finally, we input the structure code and latent codes into the pre-trained StyleGAN2 to generate the inversion. During the training stage, the middle layer distribution of StyleGAN2 and ASIT is aligned. The source image and the inversion are judged in an adversarial way.}
    \label{fig:3}
  \hfill
\end{figure*}

Furthermore, we provide extensive ablation studies to validate the theory that learning the generation of expressions improves the understanding of expressions. It is helpful for future research in this direction of FER through facial inversion. Our main contributions are as follows:
\begin{itemize}[itemsep=2pt,topsep=0pt,parsep=0pt]
    \item A novel perspective of facial expression recognition is proposed by us. We revisit this issue on whether it can acquire useful representations that improve FER performance in the image generation process.
    \item For the expression recognition problem, we design a novel method called Inversion FER. It fully fuses the structure code and the latent codes learned from facial inversion by our feature modulation module.
    \item In order to focus on facial expressions more comprehensively and effectively, we propose an image inversion method named Adversarial Style Inversion Transformer. It achieves state-of-the-art results on several facial inversion datasets and can generalize to non-face datasets.
\end{itemize}


\section{Related Work}
\label{sec:related_work}
\subsection{Facial expression recognition}
FER is a challenging task in the field of computer vision. Traditional expression recognition algorithms such as LBP \cite{bib8}, LDP \cite{bib9}, HOG \cite{bib10}, etc. recognize expressions by using manual features. However, with the availability of more and more real scene datasets, such methods have difficulty in achieving the same results as datasets in a laboratory setting.

Recently, the development of deep learning has given rise to a large number of novel expression recognition algorithms. De-expression Residue Learning (DeRF) \cite{bib28} introduces GAN to generate expressionless neutral faces to better learn the expression components in faces. DCD \cite{bib29} proposes the covariance matrix descriptor as a way to encode deep convolutional neural networks features in FER. DLN \cite{bib30} proposes a deviation learning network that uses three modules to subtract identity attributes from facial representations and calculate expression deviation features. IPER \cite{bib31} introduces adversarial learning, which considers pose variations and identity bias at the same time. RAN \cite{bib32} proposes regional attention network to capture important facial regions by fusing self-attention module and relational attention module. SCN \cite{bib42} utilizes self-attention as weights to learn the certainty in samples via models and loss functions in order to reduce the impact.

\subsection{Image inversion}
In recent years, generative adversarial networks (GAN) \cite{bib33} have become increasingly sophisticated. In particular, the generative models represented by StyleGAN \cite{bib23,bib19,bib34},they can generate realistic face images. As GAN keeps developing, numerous literatures \cite{bib35,bib36} have found that the potential space of GAN contains rich semantic information.

Zhu \etal \cite{bib37} are the first to propose using a pre-trained GAN to find a latent codes from a given image that can reconstruct the original image. In order to make the image inversion model understand the training process of GAN, Pidhorskyi \etal \cite{bib17} propose ALAE, which uses discriminators to make the latent codes simulate the real data distribution to make the reconstructed images more realistic. Xu \etal \cite{bib16} design a new feature extraction branch for the image inversion model and combine it with the AdaIN \cite{bib38} module in StyleGAN \cite{bib23} thus performing well on several tasks. During the same period, Richardson \etal \cite{bib14} also use the intermediate layer features of the model for image inversion and introduce an identity consistency loss for them to enhance the consistency between the inversion and the input image artifacts. Tov \etal \cite{bib15} discuss the effect of different styles of mapping spaces of StyleGAN on image inversion, choosing diverse potential vectors and progressive strategies for training image inversion models. To make the latent codes extract from the image inversion model more suitable for image editing, Hu \etal \cite{bib21} use a cross-attention mechanism to make the latent codes simulate the distribution of style codes in StyleGAN. Parmar \etal \cite{bib39} explore the problem that existing image inversion methods have difficulty inverting complex scene layouts and object occlusions.

\begin{figure}[t]
  \centering
   \includegraphics[width=1\linewidth]{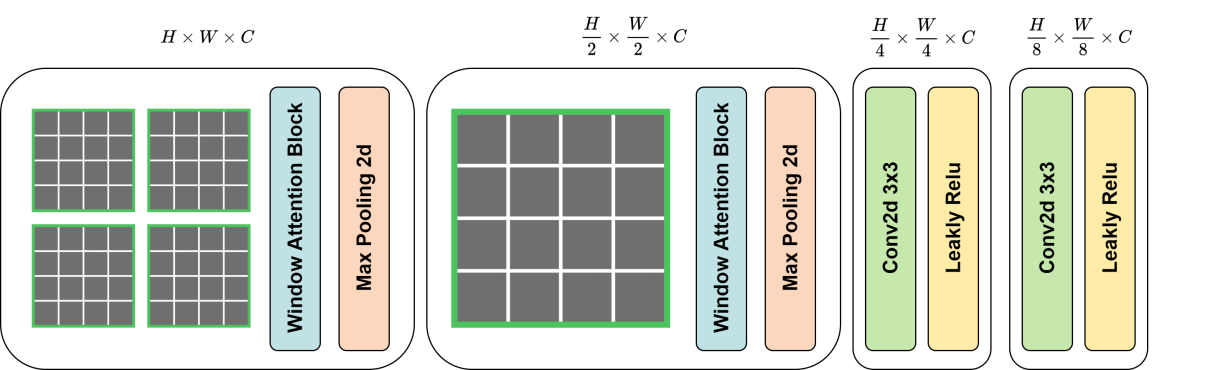}
   \caption{Our feature extraction branch architecture. The green box indicates our window. By continuously reducing the feature map size up to the window size, we finally achieve feature extraction from local to global.}
   \label{fig:4}
\end{figure}

\section{Approach}
\label{sec:approach}
To verify whether learning expression generation contributes to FER, we propose the IFER coupled with an image inversion model. Particularly, we devise a novel Adversarial Style Inversion Transformer (ASIT) towards IFER to comprehensively extract features of generated facial images.  In this section, we describe our ASIT and IFER in detail.

\subsection{Adversarial Style Inversion Transformer}
Fig. \ref{fig:3} illustrates the training process of ASIT. ASIT contains the following three main core designs: (1) To focus more comprehensively on facial features and reduce the expression recognition errors caused by focusing on only part of the face, we propose a fully transformer-based image inversion network to establish long-range dependencies; (2) We design a discriminator model in ASIT specifically to close the distance between the inversion and the high-level semantic features of the source image; (3) We introduce distributed alignment loss for ASIT to fully take advantage of the prior knowledge of StyleGAN2.

\noindent\textbf{Model Design.} We introduce GC ViT \cite{bib20} as the base model for our ASIT. There are two main reasons for using GC ViT instead of convolution network as the base model for image inversion. First, GC ViT designs a new global attention mechanism to successfully allow the image inversion model to extract global information from the input image. Second, GC ViT also uses window attention to observe local details of the image, which retains the advantages of the convolutional image inversion model. Besides, we also reconsider the feature extraction branch of the image inversion model. Our feature extraction branch mainly consists of a window attention module and a maximum pooling layer to extract image features from local to global. We set a different window size for each branch. As the pooling layer increases, the middle layer features resolution decreases and eventually reduces the window size. Our feature extraction branch is therefore able to capture intermediate vectors progressively from local to global. The overall architecture of our feature extraction branch is shown in \cref{fig:4}.

\begin{figure}[t]
  \centering
   \includegraphics[width=1\linewidth]{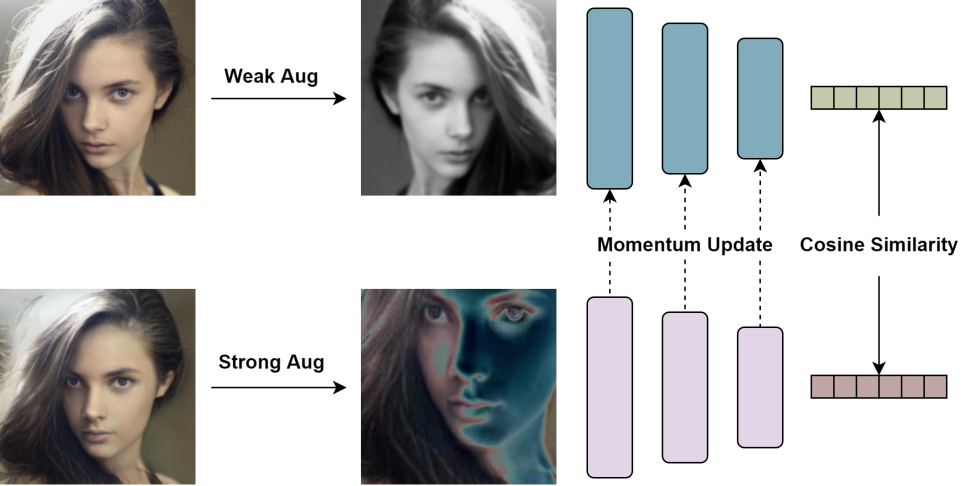}
   \caption{Our Discriminator Process. Our discriminator uses siamese architecture. The target encoder $m_{k}$ is updated by the source encoder $m_{q}$ momentum. We perform unbalanced data enhancement on a pair of input images and feed them into $m_{q}$ and $m_{k}$, respectively. The discriminator finally takes the cosine similarity between the high-level semantic information extracted by $m_{q}$ and $m_{k}$ as the output.}
   \label{fig:5}
\end{figure}

\noindent\textbf{Image Inversion Discriminator.} To better achieve the real purpose of image inversion, we design a discriminator model dedicated to image inversion. We show the process of our discriminator in \cref{fig:5}. Our image inversion discriminator mainly consists of an encoder $m_{q}$ and a momentum encoder $m_{k}$. We choose the feature extraction layer of the discriminator in StyleGAN2 as the encoder. We denote the parameters of $m_{q}$ by $\theta_{q}$ and the parameters of $m_{k}$ by $\theta_{k}$. We update $\theta_{k}$ by \cref{eq:1}.
\setlength\abovedisplayskip{0.3cm}
\setlength\belowdisplayskip{0.3cm}
\begin{equation}
\theta_{\mathrm{k}}=a \theta_{\mathrm{k}}+(1-a) \theta_{q}
\label{eq:1}
\end{equation}
where $a$ denotes the momentum update weight.

Then $m_{q}$ and $m_{k}$ extract the features of a pair of images respectively and calculate their cosine similarity as the output. Formally, we assume that the input image is $x$ and the output of ASIT can be defined as: 
\begin{equation}
\label{eq}
    y = {G}\left({E}\left({x}\right)\right)
\end{equation}
where $G$ represents the pre-trained StyleGAN2 model and $E$ denotes the encoder. Before the images are fed into the discriminator, data enhancement is performed on them. In the discriminator, we expect two identical source images to be judged as true while the inversion and the source image are judged as false. But for ASIT, we expect its inversion result and the source image to be judged as true. Through the intense confrontation between the discriminator and ASIT, we obtain the inversion, which contains similar high-level semantic information to the source image. The least squares loss \cite{bib40} is introduced as their adversarial loss. Adversarial process can be described in the form of \cref{eq:2}. 

\begin{equation}
\hspace{-3mm}
  \small
    \label{eq:2}
    \begin{aligned}
       & \min _{\mathrm{D}} L\left(D\right)=E_{x \sim P_{x}}\left(\left(D\left(x, x\right)-\mathrm{a}\right)^{2}+\left(D\left(y, x\right)-b\right)^{2}\right), \\
       & \min _{E} L\left(E\right)=E_{x \sim P_{x}}\left(\left(D\left(y, x\right)-a\right)^{2}\right), \\
       & D\left(x, x\right)=\cos \left(m_{q}\left(\phi\left(x\right)\right), m_{k}\left(\phi\left(x\right)\right)\right), \\
       & D\left(y, x\right)=\cos \left(m_{q}\left(\phi\left(y\right)\right), m_{k}\left(\phi\left(x\right)\right)\right),
    \end{aligned}
\end{equation}
where $\phi$ represents random data enhancement and $cos()$ denotes the cosine similarity calculation. Since the cosine similarity belongs to [-1,1], we set a to 1 and b to -1.

\begin{figure}[t]
  \centering
   \includegraphics[width=1\linewidth]{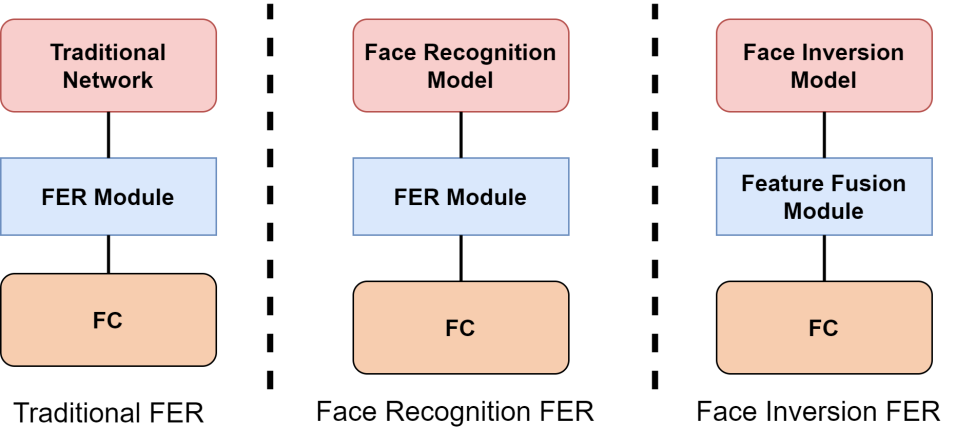}
   \caption{The main differences between our FER method and the classical FER methods. Traditional FER methods generally enhance feature extraction by designing a new network structure. Using a facial inversion model, our FER method observes faces from the perspective of facial expression generation.}
   \label{fig:6}
\end{figure}

\noindent\textbf{Distribution Alignment.} Image generation can be regarded as a process containing prior knowledge of image. We can view the image inversion model as the student and the image generation model as the teacher. If the middle layer distribution of the image inversion model is closer to that of the corresponding layer of the image generation model, the middle layer of the image inversion model can observe the same level information of the corresponding layer of the image generation model. Thereby, the image inversion model acquires prior knowledge form the image generation model by inversion learning. Different from knowledge distillation, which selects the final category distribution as the target, image inversion and image generation only yield images in the end. This result in our inability to obtain information on their distribution. To address this problem, we introduce the distribution alignment loss, which obtains the intermediate layer distribution by computing the dot product similarity between each intermediate layer feature of each image. Finally, KL divergence \cite{bib43} makes the image inversion model obtain the intermediate layer information of the image generation model. \cref{eq:3} exhibits an interpretation of our distribution alignment process.

\begin{figure}[t]
  \centering
   \includegraphics[width=1\linewidth]{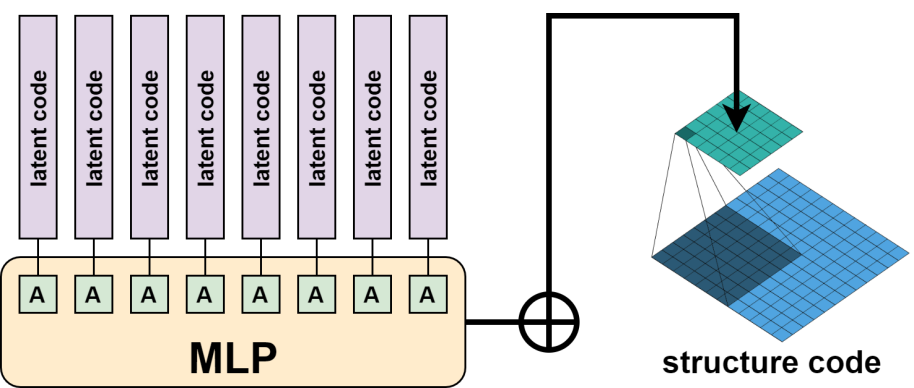}
   \caption{Our feature modulation module. We use MLP to project the latent code with different properties and sum them. A denotes a learnable affine transform. We apply the summed results to the weights associated with the modulated convolution and perform a modulation operation on the structure code.}
   \label{fig:7}
\end{figure}

\begin{equation}
    \label{eq:3}
    \begin{aligned}
    L_{\mathrm{alig}}&=\sum_{i=1}^{n} D_{K L}\left(E_{\text {dis }_{i}} \| G_{\text {dis }_{i}}\right), \\
    G_{d i s_{i}}&=\operatorname{soft} \max \left(\left\{\operatorname{sim}\left(g_{\text {feat }_{i}}(a), g_{\text {feat }_{i}}(b)\right)\right\}_{\forall a \neq b}\right), \\
    E_{d i s_{i}}&=\operatorname{soft} \max \left(\left\{\operatorname{sim}\left(e_{\text {feat }_{i}}(a), e_{\text {feat }_{i}}(b)\right)\right\}_{\forall a \neq b}\right),
    \end{aligned}
\end{equation}

Here, $a$ and $b$ denote different images in a batch, $g_{feat}$ and $e_{feat}$ are the middle layer features of the generative model and the middle layer features of the image inversion model, respectively. $sim()$ is the dot product calculation. $G_{dis_{i}}$ and $E_{dis_{i}}$ denote the distribution of the $i$-${th}$ layer of the generative model and the image inversion model. In total, we align the distribution of $n$ layers of middle layer features.

\subsection{Inversion FER}

\cref{fig:6} shows the main differences between our method and the classical FER methods. Meanwhile, IFER introduces structure code for the ASIT to better utilize the image structure information. Besides, we design a feature modulation module for IFER that successfully fuses the latent codes obtain from ASIT with structure code. 

\noindent\textbf{Structure Code.}In ASIT, the highest feature map named structure code is utilized as the initial input of the generation. The emergence of the structure code both provides structural information about images for ASIT and smoothly enables IFER to observe the high-level semantic features related to expression generation.

\noindent\textbf{Feature modulation module.} After obtaining the latent codes and the structure code from ASIT, IFER becomes concerned with how to effectively combine them. We propose a feature modulation module for IFER based on modulated convolution in StyleGAN2. \cref{fig:7} illustrates the architecture of the feature modulation module. Besides, ASIT provides multiple levels of latent codes that contain different properties of the structure code. While the structure code belongs to high-level visual features, deeper latent code is more relevant to it. Therefore, we use MLPs in the feature modulation module to map the latent codes, assigning different weights to different levels of latent codes.

\begin{table}[t]
  \centering
  \tabcolsep=0.14cm
  \scriptsize
    \begin{tabular}{cc|ccccc}
    \toprule
    \textbf{Dataset} & \textbf{Method} & \textbf{MSE}$\downarrow$ & \textbf{LPIPS}$\downarrow$ & \textbf{FID}$\downarrow$ & \textbf{SSIM}$\uparrow$ & \textbf{PSNR}$\uparrow$ \\
    \midrule
            & ALAE \cite{bib17}  & 0.182 & 0.43  & 24.86 & 0.398 & 10.657 \\
          & pSp \cite{bib14}   & 0.042 & 0.17  & 31.52 & 0.728 & 20.805 \\
     {FFHQ \cite{bib23}} & e4e \cite{bib15}  & 0.050  & 0.21 & 36.16 & 0.669 & 19.252 \\
          & ST \cite{bib21}    & 0.036 & 0.17 & 28.31 & 0.674 & 19.254 \\
          & Ours  & \textbf{0.030} & \textbf{0.14} & \textbf{19.85} & \textbf{0.755} & \textbf{21.860} \\
    \midrule
          & ALAE \cite{bib17}  & 0.41  & 0.34  & 54.34 & 0.437 & 10.900 \\
          & pSp \cite{bib14}  & 0.04  & 0.18  & 41.72 & 0.737 & 21.265 \\
{Celeba-HQ \cite{bib24}}  & e4e \cite{bib15}  & 0.05  & 0.21  & 44.58 & 0.689 & 19.670 \\
          & ST \cite{bib21}   & 0.03  & 0.16  & 39.34 & 0.768 & 21.915 \\
          & Ours  & \textbf{0.03} & \textbf{0.13} & \textbf{31.64} & \textbf{0.786} & \textbf{22.480} \\
    \bottomrule
    \end{tabular}%
    \caption{Quantitative comparison of different inversion methods on facial datasets. For an adequate comparison, we evaluate several metrics between the inversion and the input image.}
  \label{tab:1}
\end{table}%

\begin{table}[t]
  \centering
  \tabcolsep=0.14cm
  \scriptsize
    \begin{tabular}{cc|ccccc}
    \toprule
    \textbf{Dataset} & \textbf{Method} & \textbf{MSE}$\downarrow$ & \textbf{LPIPS}$\downarrow$ & \textbf{FID}$\downarrow$ & \textbf{SSIM}$\uparrow$ & \textbf{PSNR}$\uparrow$ \\
    \midrule
         & ALAE \cite{bib17}  & 0.35  & 0.63  & 49.465 & 0.309 & 11.178 \\
          & pSp \cite{bib14}  & 0.08  & 0.27  & 24.570 & 0.526 & 17.737 \\
        Church \cite{bib41}  & e4e \cite{bib15}  & 0.69  & 0.49  & 27.330 & \textbf{0.689} & \textbf{19.668} \\
          & ST \cite{bib21}   & 0.08  & 0.24  & 22.480 & 0.557 & 18.226 \\
          & Ours  & \textbf{0.06} & \textbf{0.23} & \textbf{13.624} & 0.514 & 17.343 \\
    \bottomrule
    \end{tabular}
    \caption{Quantitative comparison of different inversion methods on LSUN Church datasets.}
  \label{tab:2}
\end{table}%

\section{Experiment}
\label{sec:experiment}
To verify the effectiveness of our proposed ASIT and IFER, we conduct image inversion as well as FER experiments and compare them with currently popular methods on numerous metrics. In addition, we carefully design ablation experiments for each important component of ASIT and IFER. The ablation results demonstrate the feasibility of our proposed perspective.

\subsection{Image inversion}

\begin{figure}[t]
  \centering
   \includegraphics[width=1\linewidth]{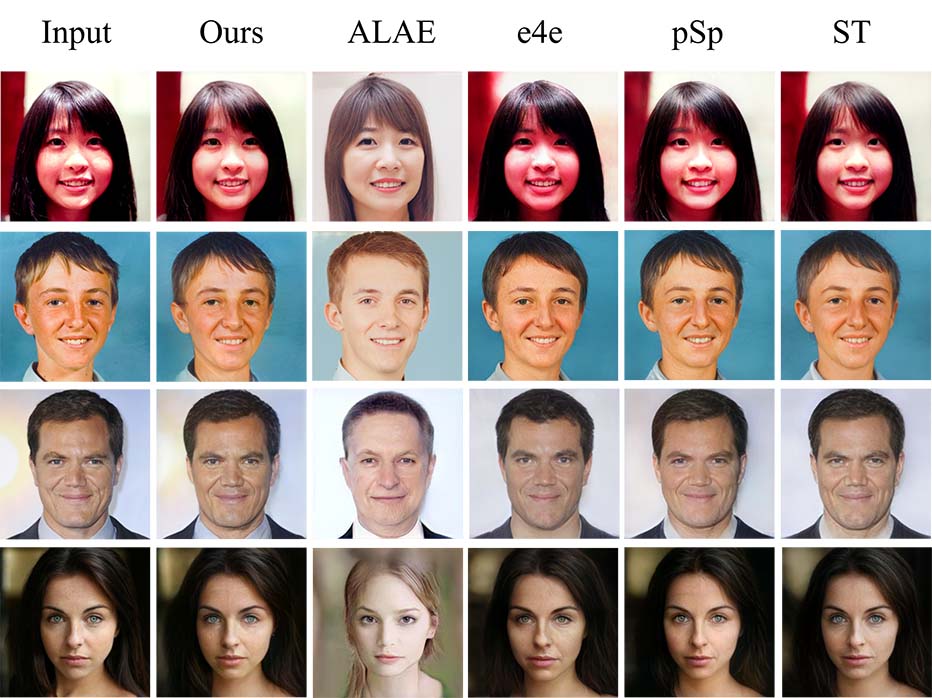}
   \caption{Comparison of the inversion results of ASIT with other image inversion algorithms on CelebA-HQ and FFHQ.}
   \label{fig:8}
\end{figure}

\begin{figure}[t]
  \centering
   \includegraphics[width=1\linewidth]{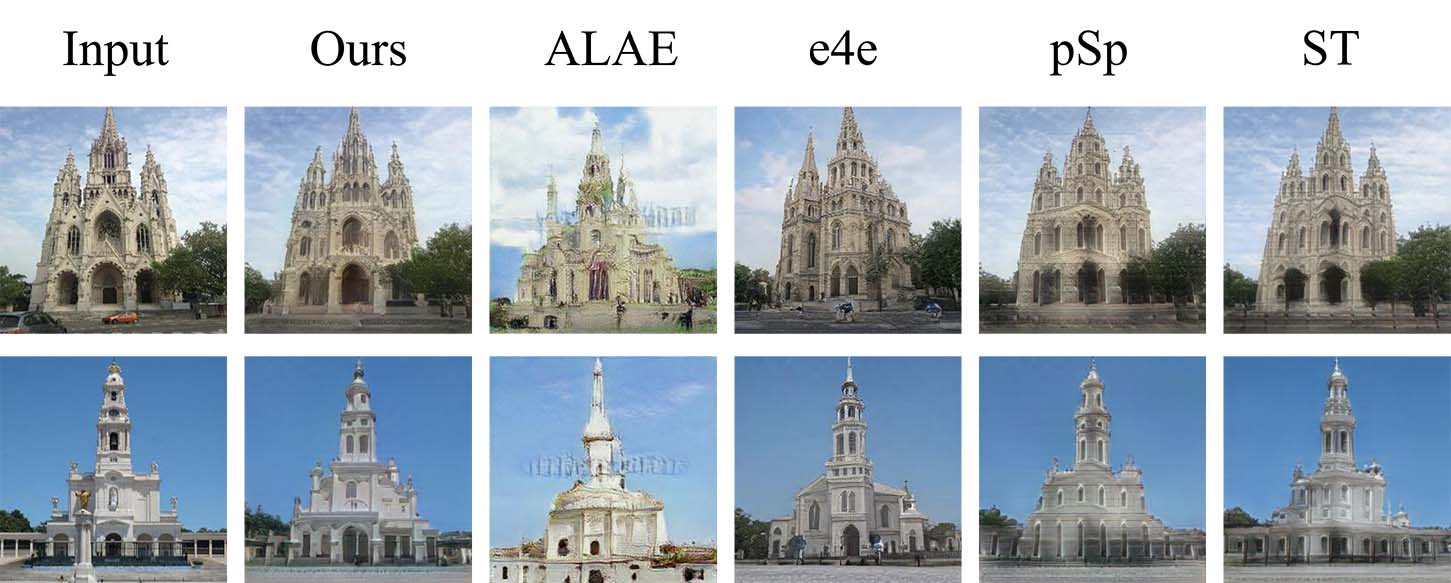}
   \caption{Visual comparison of ASIT with other image inversion models on LSUN Church.}
   \label{fig:9}
\end{figure}

\noindent\textbf{Implementation Details.} To evaluate the image inversion performance as well as the generalization capability of ASIT, we conduct experiments on facial datasets and the church \cite{bib41} dataset, respectively. All image inversion experiments are implemented based on the official version of the StyleGAN2 pre-trained model. Furthermore, all parameters in StyleGAN2 are kept fixed during the training process. For the facial datasets, we train and test on FFHQ \cite{bib23} and CelebA-HQ \cite{bib24}, respectively. For church domain, the ASIT is trained and evaluated on LSUN Church\cite{bib41}. All comparison methods are conducted in the same way. The rest of the training configuration is same as the pSp \cite{bib14}.

\noindent\textbf{Result.} We select four image inversion methods for comparison, including ALAE \cite{bib17}, e4e \cite{bib15}, pSp \cite{bib14} and ST \cite{bib21}. \cref{tab:1} and \cref{fig:8} show the image inversion performance of ASIT on the facial dataset quantitatively and qualitatively, respectively. It can be seen that our ASIT achieves the lowest MSE, LPIPS, FID and highest SSIM, PSNR metrics on the facial datasets. This indicates that the ASIT inversions are closer to the pixel values of the source images, with more similar content and higher definition. The FID metrics show that ASIT better approximates the distribution of the inversions to the source images compared to other methods. To verify the generalization capability of ASIT, we also compare it with state-of-the-art image inversion method on the LSUN Church \cite{bib41}. Qualitative and quantitative indicators are presented in Fig. \ref{fig:9} and Tab. \ref{tab:2}. Our church inversion results achieve SOTA in MSE, LPIPS and FID metrics, but the clarity of the images is slightly degraded. Besides, we also show the style mixing results for ASIT in \cref{fig:10}. See Appendix \textcolor{red}{C} for more results.

\begin{figure}[t]
  \centering
   \includegraphics[width=1\linewidth]{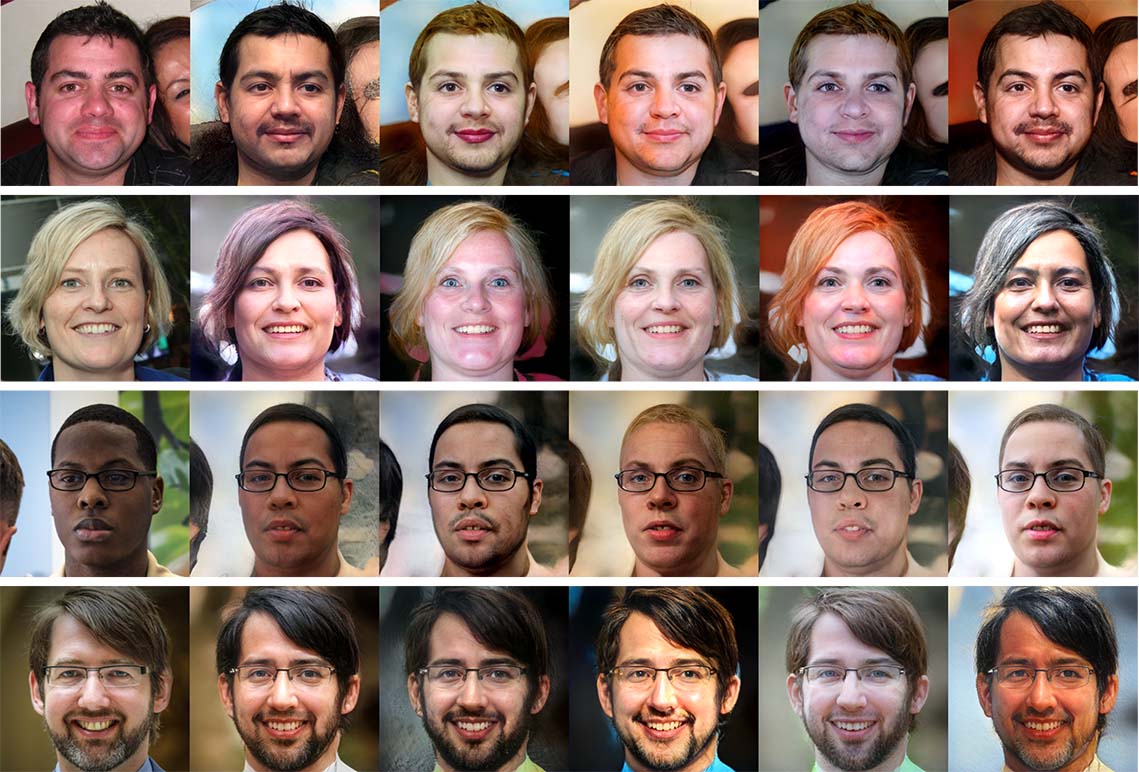}
   \caption{Visual comparison of ASIT with other image inversion models on LSUN Church.}
   \label{fig:10}
\end{figure}

\subsection{Facial Expression Recognition}
\begin{figure*}[t]
  \centering
   \includegraphics[width=1\linewidth]{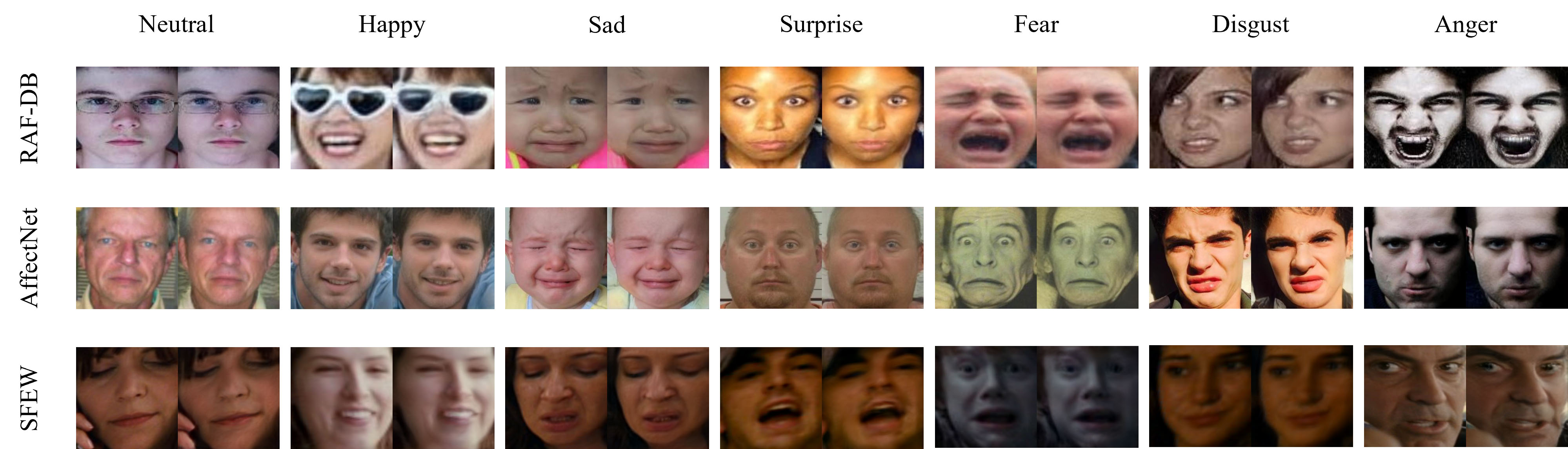}
   \caption{Results of ASIT fine-tuning on three FER standard datasets. For the 7 different types of expressions, ASIT has a good grasp of how they are generated.}
   \label{fig:11}
\end{figure*}

\noindent\textbf{Implementation Details.} We fine-tune the trained facial inversion model on the FER training sets. \cref{fig:11} shows the results of our fine-tuning. We combine the fine-tuned facial inversion model with the feature modulation module as IFER, which experiments on the RAF-DB \cite{bib25}, SFEW \cite{bib27} and AffecNet \cite{bib26}. See Appendix \textcolor{red}{A} for detailed training configurations.

\noindent\textbf{Results.}  Except the classical FER methods including RAN \cite{bib32}, SCN \cite{bib42}, DLN \cite{bib30}, DCD \cite{bib29}, IPER \cite{bib31} are selected for comparison, we apply the image inversion models to the expression recognition task as well to verify the effectiveness of IFER. All their configurations are kept the same as IFER. \cref{tab:3} shows the results of comparing IFER with other methods on each of the three datasets. This demonstrates that our proposed ASIT is more suitable for the FER in comparison to other image inversion models. Moreover, our IFER achieves competitive results compared with FER methods of recent years. This indicates that the FER method based on image inversion is worth investigating.

\subsection{Ablation Study}
\noindent\textbf{ASIT ablation experiments}. To verify the validity of our proposed innovation points in ASIT, we ablate Feature extraction branch, Image inversion discriminator and Distribution alignment loss in the FFHQ, respectively. The statistics in \cref{tab:4} show that our innovation points are indispensable for the successful facial inversion of ASIT.

\begin{table}[t]
  \centering
  \footnotesize
    \begin{tabular}{ccc|cc}
    \toprule
    \textbf{Method} & \textbf{RAF-DB} & \textbf{AffectNet} & \textbf{Method} & \textbf{SFEW} \\
    \midrule
    RAN \cite{bib32}   & 86.9  & 59.5  & RAN \cite{bib32}  & \textbf{55.1} \\
    SCN \cite{bib42}  & 87.03 & 60.23 & DCD \cite{bib29}  & 49.18 \\
    DLN \cite{bib30}  & 86.4  & \textbf{63.7} & IPER \cite{bib31} & 54.19 \\ \hline
    ALAE+FER & 80.48 & 51.85 & ALAE+FER & 41.37 \\
    pSp+FER & 86.27 & 53.11 & pSp+FER & 46.64 \\
    e4e+FER & 87.06 & 57.74 & e4e+FER & 36.66 \\
    ST+FER & 81.323 & 57.69 & ST+FER & 37.12 \\
    \textbf{IFER(Ours)} & \textbf{87.23} & 63.58 & \textbf{IFER(Ours)} & 54.37 \\
    \bottomrule
    \end{tabular}%
  \caption{Performance comparison($\%$) with FER methods based on image inversion model and conventional FER methods on RAF-DB, SFEW and AffectNet.}
  \label{tab:3}%
\end{table}

\begin{table}[t]
  \centering
    \tabcolsep=0.15cm
    \footnotesize
    \begin{tabular}{cccccccc}
    \toprule
    \textbf{DAL} & \textbf{WAB} & \textbf{IID} & \textbf{MSE}$\downarrow$ & \textbf{LPIPS}$\downarrow$ & \textbf{FID}$\downarrow$ & \textbf{SSIM}$\uparrow$ & \textbf{PSNR}$\uparrow$ \\
    \midrule
    \checkmark     &                & \checkmark     & 0.04  & 0.15  & 21.12 & 0.743 & 21.31 \\
    \checkmark     & \checkmark     & S              & 0.03  & 0.15  & 21.57 & 0.744 & 21.35 \\
    \checkmark     & \checkmark     &                & 0.04  & 0.16  & 23.89 & 0.738 & 21.10 \\
                   & \checkmark     & \checkmark     & 0.04  & 0.14  & 20.74 & 0.742 & 21.28 \\
    \midrule
    \checkmark     & \checkmark     & \checkmark     & \textbf{0.03} & \textbf{0.14} & \textbf{19.85} & \textbf{0.755} & \textbf{21.86} \\
    \bottomrule
    \end{tabular}%
    \caption{ASIT ablation experiments on the FFHQ dataset. The ASIT ablation configuration follows the ASIT training configuration on FFHQ. S stands for the discriminator in StyleGAN2 \cite{bib23}.}
 \label{tab:4}
\end{table}

Compared with traditional convolutional feature extraction branch, our window attention branch(\textbf{WAB}) achieves better results. The rest of the experimental design keep the same setting. Besides, to better show the applicability of our image inversion discriminator(\textbf{IID}) for the image inversion task, we replace the image inversion discriminator of ASIT with the discriminator in StyleGAN2. The Results show that our image inversion discriminator contributes to all the metrics to some degree, which illustrates that our image inversion discriminator is more suitable for image inversion tasks. We further investigate the effectiveness of the distribution alignment loss(\textbf{DAL}) in ASIT. It can be seen that the distribution alignment loss is particularly important for the content and distribution similarity between the two images. However, it is not very helpful for other indicators.

\begin{table*}[t]
\vspace{-.2em}
\centering
\subfloat[
\textbf{Structure code}. Without structure code, IFER performance is degraded.
\label{tab:5a}
]{
    \centering
    \begin{minipage}{0.29\linewidth}{\begin{center}
    \tablestyle{4pt}{1.05}
    \begin{tabular}{cc}
    Method & Acc  \\
    \toprule[0.2mm]
    structure code & \textbf{87.32}  \\
    w/o structure code & 86.69  \\
    \multicolumn{2}{c}{~}\\
    \end{tabular}
    \end{center}}\end{minipage}
}
\hspace{1em}
\subfloat[
\textbf{Feature modulation module}. Use our modules or only one of these features.
\label{tab:5b}
]
    {
    \begin{minipage}{0.29\linewidth}{\begin{center}
    \tablestyle{4pt}{1.05}
        \begin{tabular}{cc}
        Method & Acc \\
        \toprule[0.2mm]
        feature modulation  & \textbf{87.32} \\
        middle layer features & 86.71 \\
        structure code & 86.55 \\
        \end{tabular}%
    \end{center}}\end{minipage}
    }
\hspace{1em}
\subfloat[
\textbf{FER with pre-training model}. The pre-training model using facial inversion brings a significant effect improvement.
\label{tab:5c}
]
    {
    \begin{minipage}{0.29\linewidth}{\begin{center}
    \tablestyle{4pt}{1.05}
    \begin{tabular}{cc}
        Method & Acc \\
        \toprule[0.2mm]
        w/ pre-training model & \textbf{87.32} \\
        w/o pre-training model & 83.8 \\
    \multicolumn{2}{c}{~}\\
    \end{tabular}
    \end{center}}\end{minipage}
    }
\hspace{1em}
\caption{ \textbf{IFER ablation experiments} on the RAF-DB. We report the Top-1 accuracy(\%). }
\label{tab:5} 
\end{table*}

\noindent\textbf{IFER ablation experiments}. We validate the IFER key components on the RAF-DB. First, we confirm the validity of the high-level visual information contained in the structure code. Second, we check whether the feature modulation module can effectively combine feature vectors from middle layers with high-level visual features. Finally, we examine whether a pre-trained image inversion model is utilized in our proposed research direction of expression analysis based on expression generation.

Structure code. In the feature modulation module, we use the middle layer features themselves to replace the structure code. \cref{tab:5a} shows the ablation results for the structure encoding. The experimental results illustrate that our proposed structural encoding contains high-level visual information that is beneficial to FER.

Feature modulation module. If the feature modulation module is not applied, only the middle layer features or structure code are employed to evaluate the effect of FER. The experimental results presented in \cref{tab:5b} demonstrate that the FER results obtained using only middle layer features or only structural encoding are unsatisfactory.

FER with facial inversion model. \cref{tab:5c} implicates that the pre-training model improves the accuracy of FER by 5\%. Therefore, it is feasible to analyze expressions from the perspective of facial expression generation.

\section{Conclusion}
In this paper, we propose a new point for FER, \ie , acquire useful representations in the image generation process. To this end, we propose a novel FER method combining image inversion models called IFER and achieve competitive results on several FER datasets. However, current image inversion models do not provide a global understanding of facial features and may result in incorrect expression understanding. Thereby, we design a new image inversion model, Adversarial Style Inversion Transformer. Our ASIT achieves state-of-the-art performance on multiple datasets of facial inversion. And by further ablation studies of IFER, the expression generation feature show effective for FER. This suggests that the combination of FER and image inversion is a worthwhile direction for future research.

\section*{Acknowledge} This work was supported by Public-welfare Technology Application Research of Zhejiang Province in China under Grant LGG22F020032, and Key Research and Development Project of Zhejiang Province in China under Grant 2021C03137.

\clearpage

{\small
\bibliographystyle{ieee_fullname}
\bibliography{reference}

}

\clearpage
\appendix

{\LARGE\noindent\textbf{Appendix}}

\vspace{0.5cm}
\section{Implement Details}

For Image Inversion stage, a StyleGAN2 generator with pre-trained on the FFHQ as the decoder of our ASIT. We use Ranger as our optimizer and the learning rate is set to $1\times10^{-4}$. The iterations is 500000 with a batch size of 8. We resize the input images to $256\times256$ resolution before feeding them to ASIT. Thus, the input image size and the inversion size are kept the same.
The loss function in pSp is defined as:

\setlength\abovedisplayskip{0.3cm}
\setlength\belowdisplayskip{0.3cm}
\begin{equation}
\footnotesize
    \mathcal{L}(x)=\lambda_{1} \mathcal{L}_{2}(x)+\lambda_{2} \mathcal{L}_{L P I P S}(x)+\lambda_{3} \mathcal{L}_{I D}(x)+\lambda_{4} \mathcal{L}_{\text {reg }}(x)
\end{equation}
where $\lambda_{1}$, $\lambda_{2}$, $\lambda_{3}$, $\lambda_{4}$ are constants defining the loss weights. The $\mathcal{L}_{2}$ represents the pixel-wise loss, $\mathcal{L}_{LPIPS}$ represents the LPIPS loss, $\mathcal{L}_{ID}$ represents the loss of identity consistency and $\mathcal{L}_{reg}$ represents the regularization loss about latent style vectors.

On the basis of pSp, we add distribution alignment loss $\mathcal{L}_{alig}$  to close the distribution of middle layer between encoder and generator. The loss function in ASIT can be defined as:

We set $\lambda_{1}=0.8$, $\lambda_{2}=1$, $\lambda_{3}=0.1$ and $\lambda_{4}=1\times10^{-4}$ for facial datasets. $\lambda_{adv}$ is the generated adversarial loss between our image inversion discriminator and ASIT, and we default its weight $\lambda_{5}$ to $1\times10^{-3}$. For the church domain, we use the feature consistency loss $\mathcal{L}_{moco}$ based on the pre-trained MoCo encoder to replace the identity consistency loss. For $\mathcal{L}_{moco}$, we set its weight to 2 and keep the rest of the settings unchanged.

For FER stage, we use FFHQ pretrained ASIT as our IFER backbone. Before training the FER network, we fine-tune the image inversion model on the FER training sets with 20K iterations. We use AdamW as our optimizer and the learning rate is set to $1\times10^{-3}$. The iterations are 320 with a batch size of 64. All experiments are implemented on 4 NVIDIA RTX 3090 GPUs.

\section{Dataset Details}
For Image Inversion experiments, we conduct on facial domain(FFHQ and CelebA-HQ) and church domain(LSUN Church).

\noindent\textbf{CelebA-HQ} The CelebA-HQ contains 30K face images with a resolution of $1024\times1024$. Images are split to 24K training samples and 6K test samples.

\noindent\textbf{FFHQ} 70K face images are split into 60K training samples and 10K test samples, which the resolution is $1024\times1024$.

\noindent\textbf{LSUN Church} We randomly select 10K images for train and 400 images for evaluation. The images in the dataset do not have a standard size. We will resize the resolution of the images to $256\times256$ before they are fed into the model.

For the FER experiment, we choose RAF-DB, SFEW and AffecNet as our experimental datasets.

\noindent\textbf{RAF-DB} It is a real-world expression dataset that contains 29672 real-world facial images. The dataset is divided into seven classes of expressions. The images in this database are highly variable in terms of subjects' age, gender and race, head posture, lighting conditions, occlusions, and post-processing operations.

\noindent\textbf{SFEW} The Static Facial Expressions in the Wild (SFEW) dataset is a dataset for facial expression recognition. The most commonly used version is divided into three sets: Train (958 samples), Val (436 samples) and Test (372 samples).

\noindent\textbf{AffectNet} It is a large facial expression dataset with around 0.4 million images manually labeled for the presence of seven facial expressions. In addition, the validation set consists of 3500 images, 500 for each category. For a fair comparison, 280,000 images in the AffecNet are selected for training and 4,000 images for validation.

\section{More Results}
\begin{figure*}[t]
  \centering
   \includegraphics[width=1\linewidth]{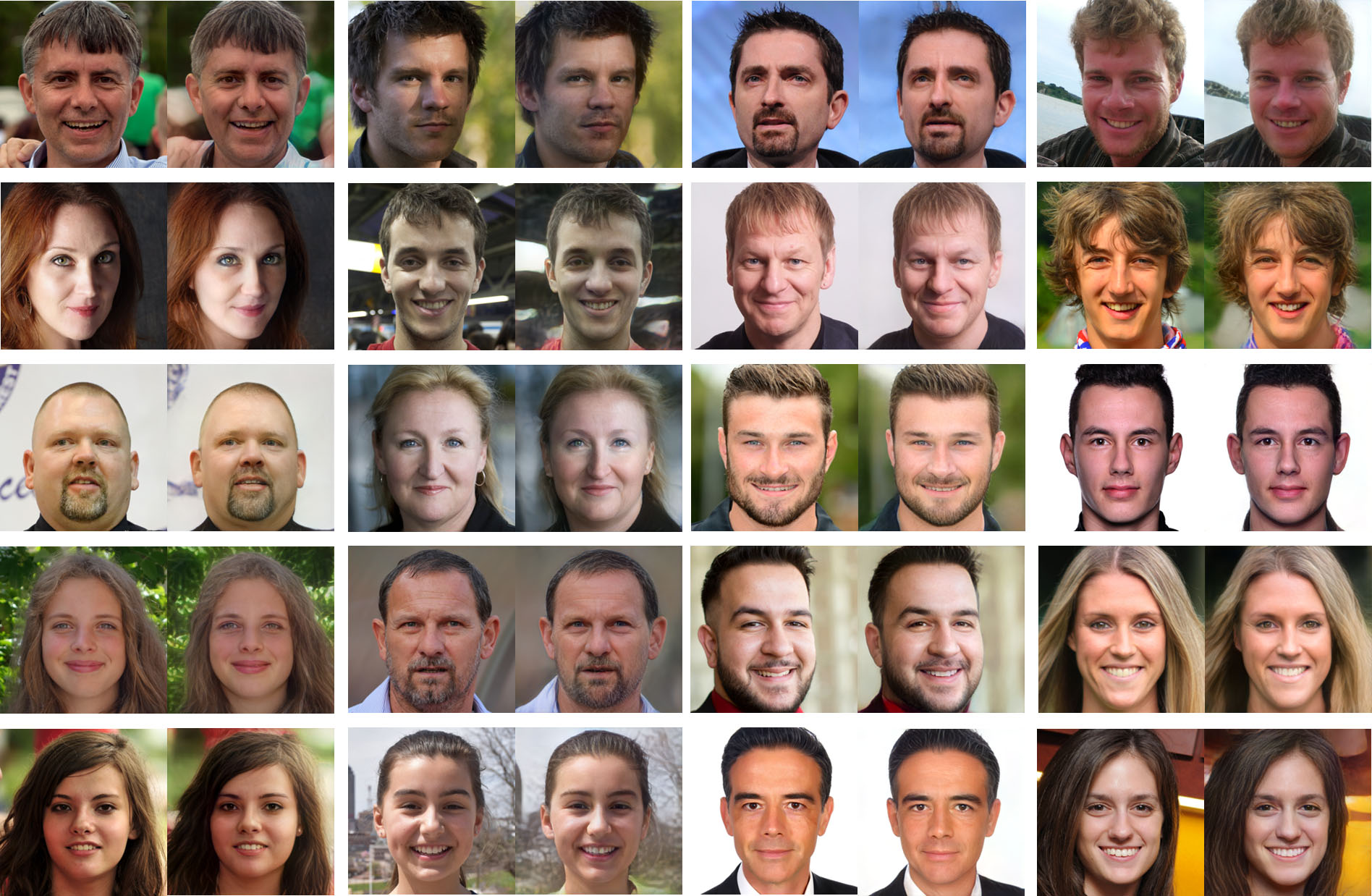}
   \caption{ More results of FFHQ inversion. For each group, the source image is on the left and the inversion is on the right.}
   \label{fig:12}
\end{figure*}

\begin{figure*}[t]
  \centering
   \includegraphics[width=1\linewidth]{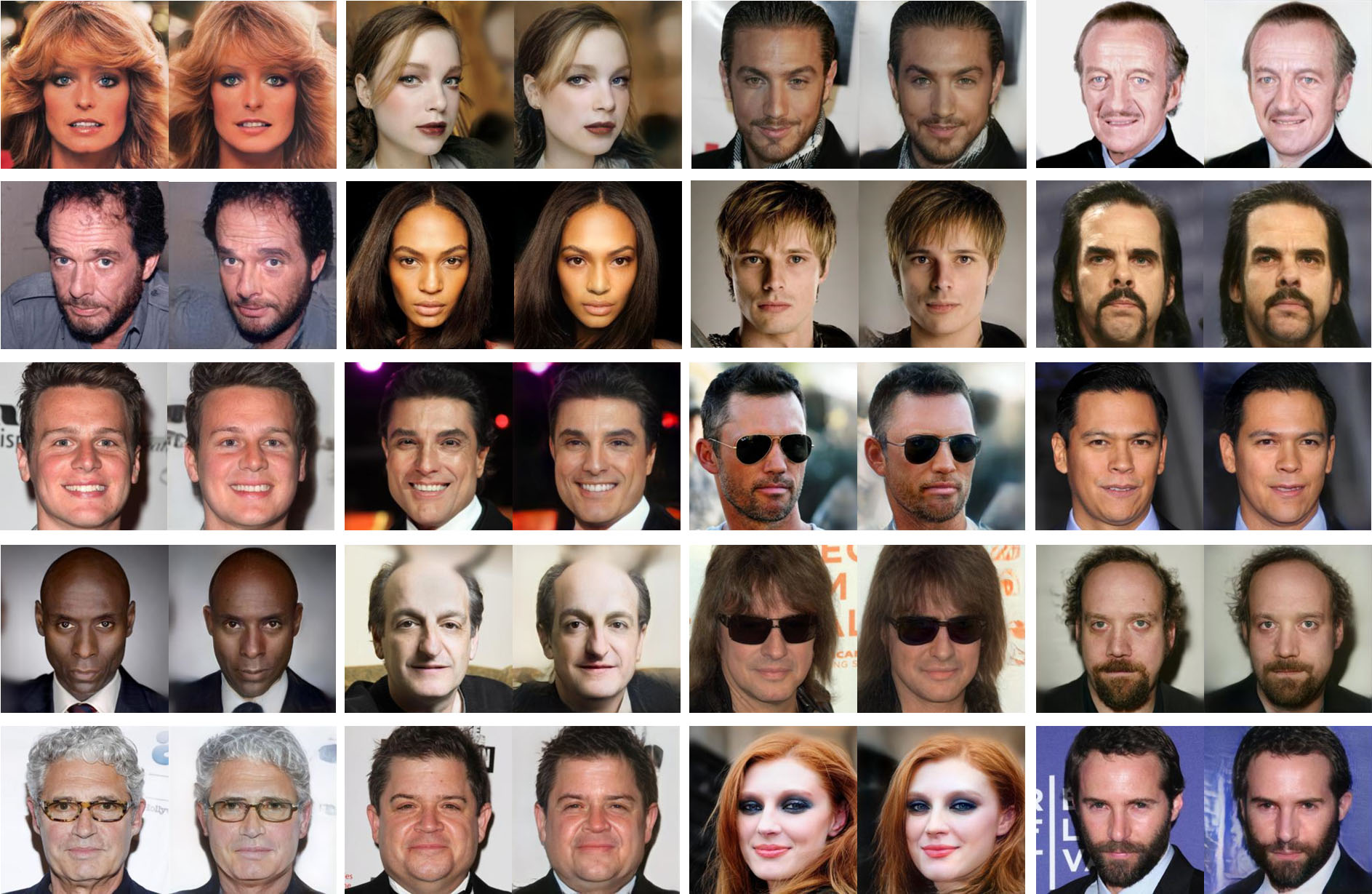}
   \caption{More results of the CelebA-HQ inversion. For each group, we show the source image on the left end and the inversion on the right.}
   \label{fig:13}
\end{figure*}

\begin{figure*}[t]
  \centering
   \includegraphics[width=1\linewidth]{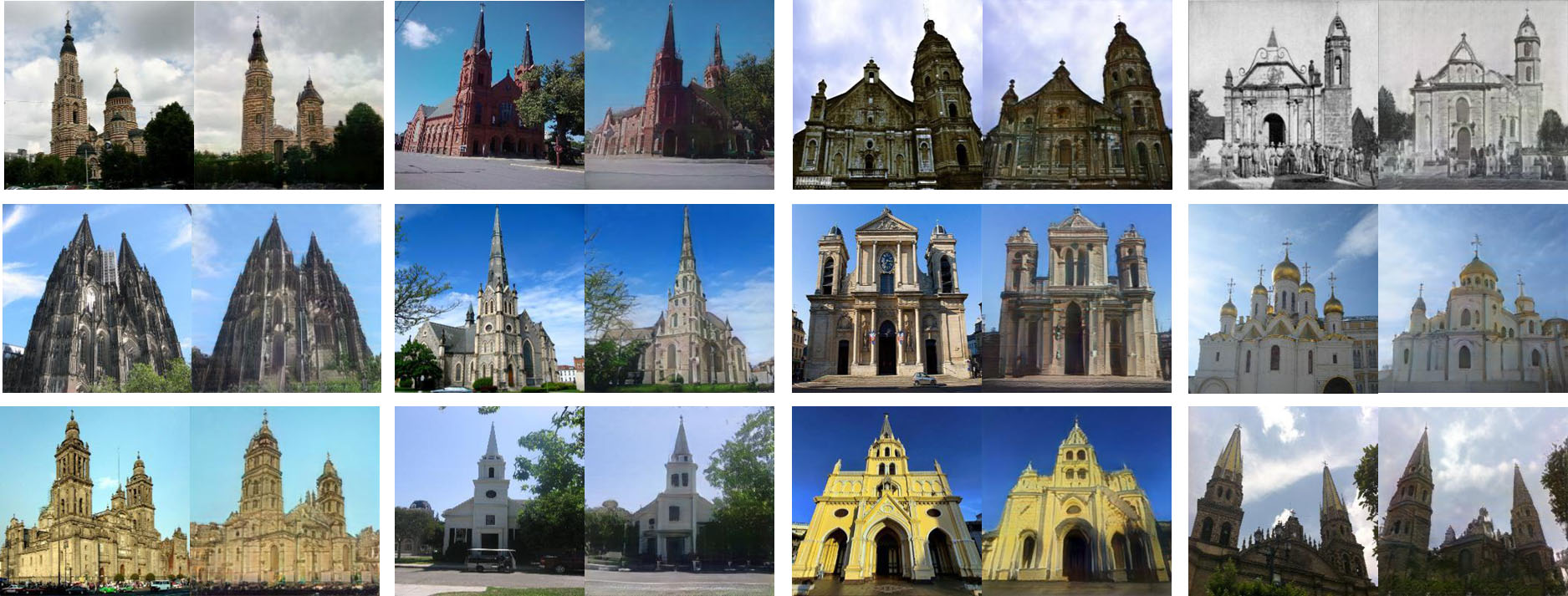}
   \caption{More results of LSUN Church inversion. For each group, the left side is the input image and the right side is the inversion.}
   \label{fig:14}
\end{figure*}

\begin{figure*}[t]
  \centering
   \includegraphics[width=1\linewidth]{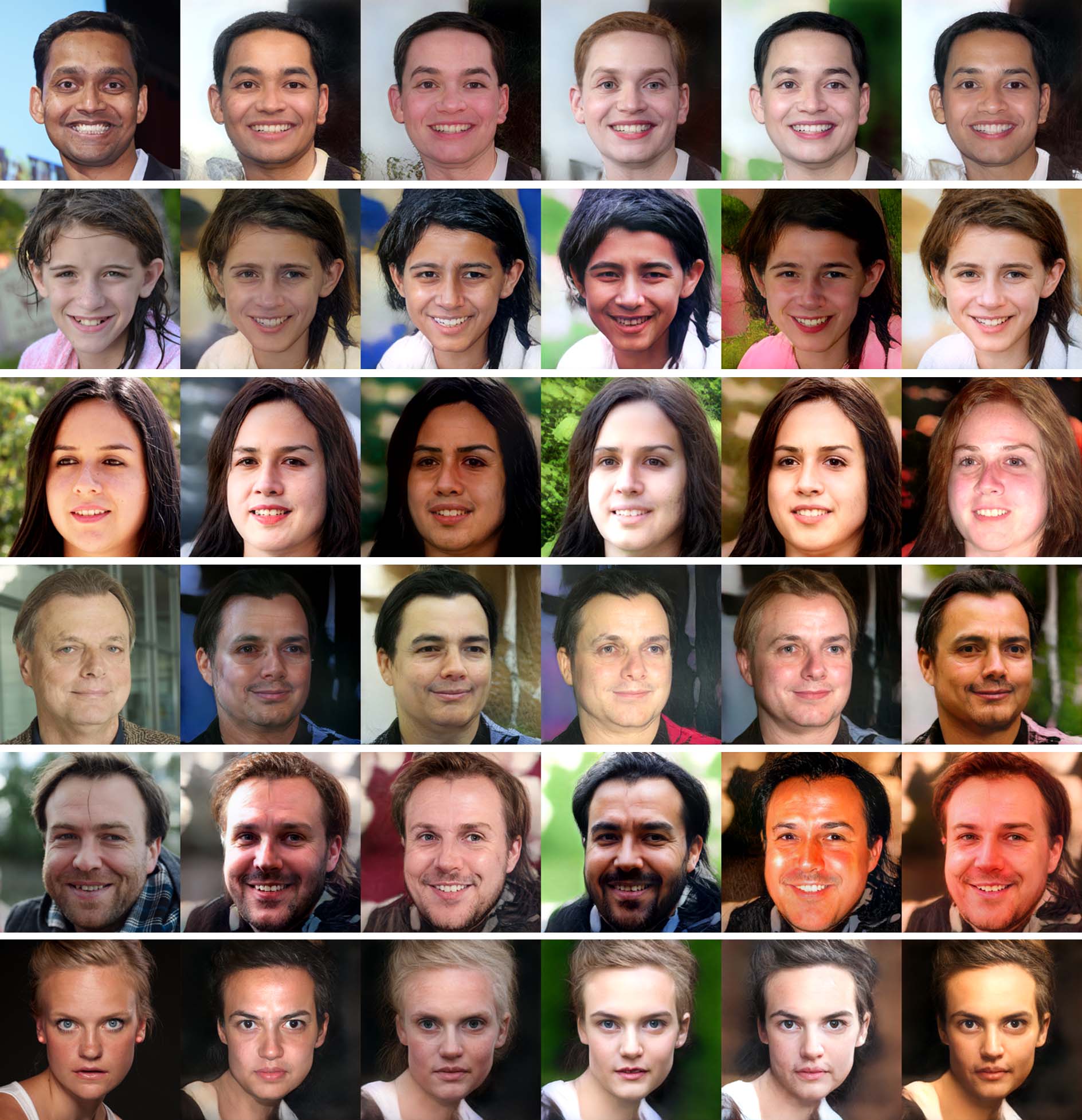}
   \caption{More style mixing results from ASIT. The left side is the input image, and the right side is the result of style mixing of the input image.}
   \label{fig:15}
\end{figure*}

\begin{figure*}[t]
  \centering
   \includegraphics[width=0.8\linewidth]{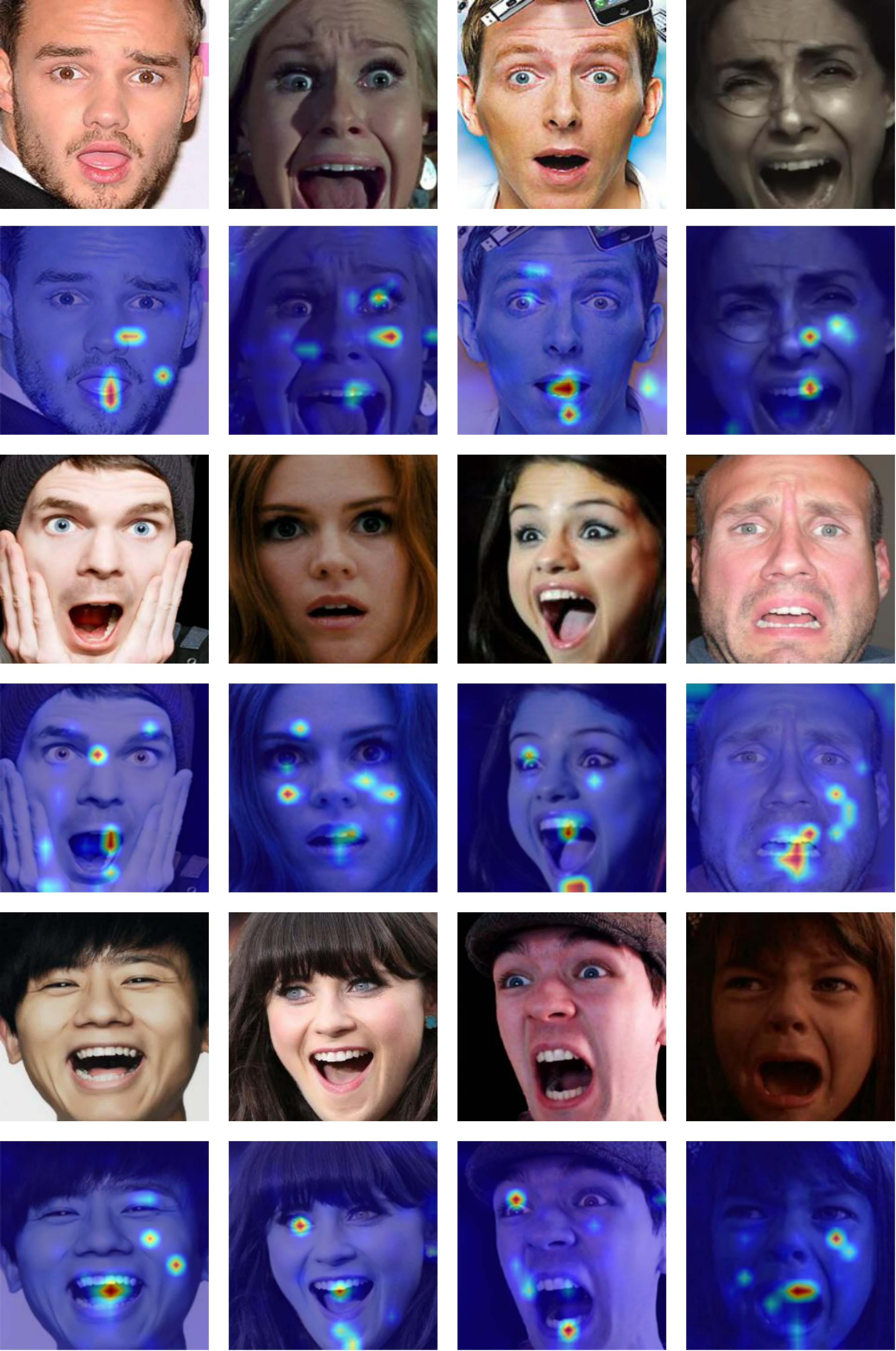}
   \caption{More results of ASIT's attention visualization on the expression dataset. We use ASIT for attention visualization of images on AffecNet.}
   \label{fig:16}
\end{figure*}

In this section, we provide more results of facial inversion and church inversion in \cref{fig:12}, \cref{fig:13} and \cref{fig:14}. Additional, \cref{fig:15} shows more style mixing results. \cref{fig:16} shows more results of the attention visualization.

\end{document}